\documentclass[10pt,twocolumn,letterpaper]{article}

\usepackage[accsupp]{axessibility}  % Improves PDF readability for those with disabilities.
\usepackage{iccv}
\usepackage{times}
\usepackage{epsfig}
\usepackage{graphicx}
\usepackage{amsmath}
\usepackage{amssymb}
\usepackage{graphicx}
\usepackage{amsmath}
\usepackage{amssymb}
\usepackage{booktabs}
\usepackage{stfloats}

\usepackage{overpic}
\usepackage[ruled,linesnumbered]{algorithm2e}

\usepackage{times}
\usepackage{epsfig}
\usepackage{graphicx}
\usepackage{amsmath}
\usepackage{mathrsfs}
\usepackage{amssymb}
\usepackage{multirow}
\usepackage{graphics}
\usepackage{threeparttable}
\usepackage{color}
\usepackage{framed}
\definecolor{shadecolor}{rgb}{.99,.91,.95}
\usepackage[normalem]{ulem}
\usepackage{float}
\usepackage{amsfonts}
\usepackage{bm}
\usepackage{bbm}
\usepackage{enumitem}
\usepackage[numbers]{natbib}
\usepackage{subcaption}
\usepackage{array}
\usepackage{colortbl}
\usepackage{pifont}

\usepackage{mathtools}
\usepackage{siunitx}
\usepackage[nopar]{lipsum}
\usepackage{listings}
\usepackage[export]{adjustbox}
\usepackage{verbatim}
\usepackage{colortbl}
\usepackage[colorlinks]{hyperref}
\usepackage{bbding}
\usepackage{pifont}
\usepackage{wasysym}
\usepackage{amssymb}

\usepackage[ruled,linesnumbered]{algorithm2e}
\usepackage[dvipsnames]{xcolor}
\newcommand{\stdvu}[1]{\scriptsize{\color{darkgray}(#1)} {\color{ForestGreen}$\uparrow$}}
\newcommand{\stdvd}[1]{\scriptsize{\color{darkgray}(#1)} {\color{red}$\downarrow$}}
\newcommand{\stdvw}[1]{\scriptsize{\color{darkgray}(#1)} {\color{white}$\downarrow$}}
\newcommand{\stdvno}[1]{\scriptsize{\color{darkgray}(#1)} {\color{mypink}$\downarrow$}}
\definecolor{mypink}{rgb}{.99,.91,.95}
\usepackage{etoolbox}
\makeatletter
\let\@algcomment\relax

\newcommand\algcomment[1]{\def\@algcomment{\footnotesize#1}}
\renewcommand\fs@ruled{\def\@fs@cfont{\bfseries}\let\@fs@capt\floatc@ruled
	\def\@fs@pre{\hrule height.8pt depth0pt \kern2pt}%
	\def\@fs@post{}%
	\def\@fs@mid{\kern2pt\hrule\kern2pt}%
	\let\@fs@iftopcapt\iftrue}
\makeatother
\usepackage{tabulary}
\newcolumntype{I}{!{\vrule width 1pt}}

\newcolumntype{x}[1]{>{\centering\arraybackslash}p{#1pt}}
\newcolumntype{y}[1]{>{\raggedright\arraybackslash}p{#1pt}}
\newcolumntype{z}[1]{>{\raggedleft\arraybackslash}p{#1pt}}
\newlength\savewidth

\iccvfinalcopy % *** Uncomment this line for the final submission

\def\iccvPaperID{1619} % *** Enter the ICCV Paper ID here

\ificcvfinal\fi

\begin{document}

%%%%%%%%% TITLE
\title{Towards Instance-adaptive Inference for Federated Learning}

\author{Chun-Mei Feng$^1$\quad Kai Yu$^{1*}$\quad Nian Liu$^{2}$\quad Xinxing Xu$^{1}$\thanks{Corresponding author.} \quad Salman Khan$^{2,3}$ \quad Wangmeng Zuo$^4$ \\
$^1$Institute of High Performance Computing (IHPC), \\Agency for Science, Technology and Research (A*STAR), Singapore\\
$^2$Mohamed bin Zayed University of Artificial Intelligence (MBZUAI), UAE\\
$^3$Australian National University, Canberra ACT, Australia\\
$^4$Harbin Institute of Technology, Harbin, China\\
{\tt\small fengcm.ai@gmail.com; yu\underline{ }kai@ihpc.a-star.edu.sg; xuxinx@ihpc.a-star.edu.sg}
\\ {\small {\url{https://github.com/chunmeifeng/FedIns}}
}
%\thanks{Corresponding author.}
%\thanks{Corresponding author.}
% \and
}
\maketitle
\ificcvfinal\thispagestyle{empty}\fi

\begin{abstract}
Federated learning (FL) is a distributed learning paradigm that enables multiple clients to learn a powerful global model by aggregating local training. 
However, the performance of the global model is often hampered by non-i.i.d. distribution among the clients, requiring extensive efforts to mitigate inter-client data heterogeneity.
%In our research, we identify a key problem of FL: there is not only inter-client data heterogeneity but also intra-client heterogeneity {\color{red}caused by the complex real-world environment}, which also significantly degrades FL performance. 
Going beyond inter-client data heterogeneity, we note that intra-client heterogeneity can also be observed on complex real-world data and seriously deteriorate FL performance.
% Our research identifies a critical issue in FL: not only is there inter-client data heterogeneity, but there is also intra-client heterogeneity due to the complex real-world data, thereby significantly reducing FL performance.
%To solve this problem, we propose a new algorithm based on a parameter-efficient fine-tuning method called `scale and shift deep features' (SSF). 
In this paper, we present a novel FL algorithm, i.e., FedIns, to handle intra-client data heterogeneity by enabling instance-adaptive inference in the FL framework.
% To address this problem, we propose a novel algorithm called FedIns based on a parameter-efficient fine-tuning method known as `scale and shift deep features' (SSF).
% FedIns makes instance-adaptive models feasible for FL by modulating a specific SSF for each input image based on a powerful pre-trained model and dynamically guiding the local process to relieve intra-client data heterogeneity.
Instead of huge instance-adaptive models, we resort to a parameter-efficient fine-tuning method, i.e., scale and shift deep features (SSF), upon a pre-trained model.
%Our approach, termed FedIns, makes instance-adaptive models feasible for FL by modulating a specific SSF for each input image based on a powerful pre-trained model and dynamically guides the local to relieve intra-client data heterogeneity.
% To solve this problem, we propose a new algorithm upon parameter-efficient fine-tuning method SSF, termed FedIns, to make instance-adaptive models feasible for FL, which modulates a specific SSF for each image input based on a powerful pre-trained model and dynamically guides the local to relieve intra-client data heterogeneity. 
Specifically, we first train an SSF pool for each client, and aggregate these SSF pools on the server side, thus still maintaining a low communication cost.
% then dynamically select relevant SSF subsets based on instance-wise input features, and finally, only aggregate these SSF pools on the server side, thus significantly reducing the communication cost. 
To enable instance-adaptive inference, for a given instance, we dynamically find the best-matched SSF subsets from the pool and aggregate them to generate an adaptive SSF specified for the instance, thereby reducing the intra-client as well as the inter-client heterogeneity. 
% Our federated SSF pool adapts to each instance, dynamically guiding clients to process the corresponding input, thereby reducing both inter- and intra-client heterogeneity. 
Extensive experiments show that our FedIns outperforms state-of-the-art FL algorithms, e.g., a $6.64\%$ improvement against the top-performing method with less than $15\%$ communication cost on Tiny-ImageNet. 
% Our code and models will be publicly released. 
%The code will be released upon acceptance.

\end{abstract}
%%%%%%%%% BODY TEXT
\section{Introduction}\label{sec:intro}
The availability of large-scale data dramatically promotes the development of deep learning models. However, as these abundant data tend to be distributed across many devices due to logistical and privacy concerns, decentralized training is often required to train the deep neural network~\cite{al2020generalizing}. As a promising distributed learning paradigm, federated learning (FL) can train global models in a distributed manner without sharing local private data~\cite{li2020federated,mcmahan2017communication,lim2020federated}. During this process, each client first trains a local model on their private data and then sends the model parameters to the server for aggregation and distribution back to the client~\cite{mcmahan2017communication}.

\begin{figure}[t]
	% \vspace{-2pt}
	\begin{center}
		\includegraphics[width=0.98\linewidth]{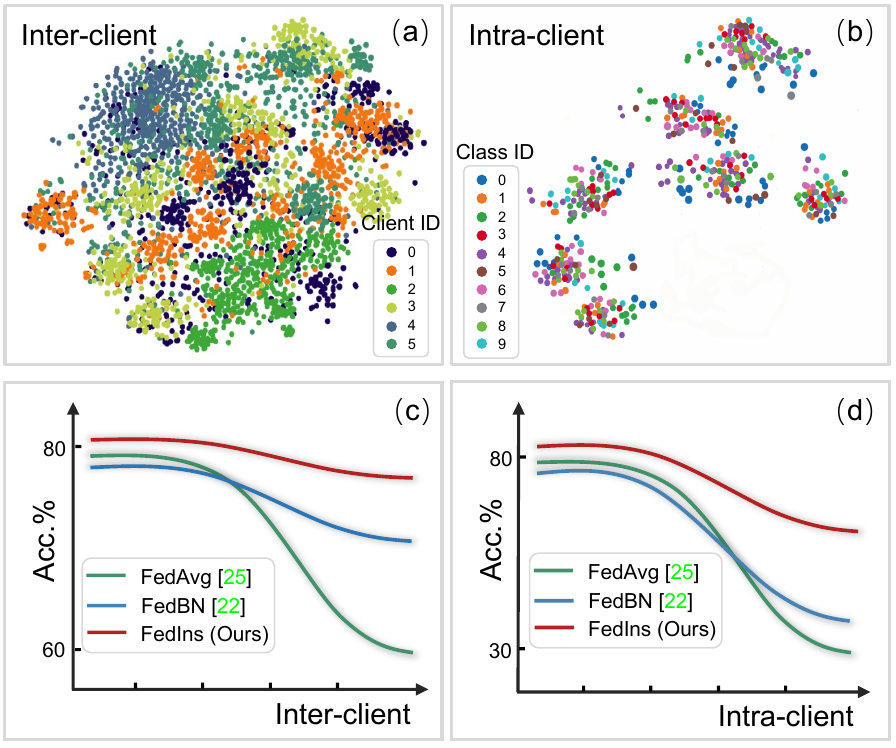}
	\end{center}
	\vspace{-14pt}
	\captionsetup{font=small}
	\caption{\textbf{Illustration} of \textbf{inter-} and \textbf{intra-client} data heterogeneity with t-SNE visualizations (see (a) and (b)) on DomainNet and their effect on accuracy of different FL algorithms (see (c) and (d)). Please refer to the \emph{Suppl.} for details on the settings to increase data heterogeneity. One can see that both inter- and intra-client data heterogeneity degrades FL performance. While FedBN~\cite{li2021fedbn} is able to alleviate  the effect of inter-client data heterogeneity, both FedAvg~\cite{mcmahan2017communication} and FedBN~\cite{li2021fedbn} are limited in handling intra-client data heterogeneity.}
	\vspace{-13pt}
	\label{fig:1}
\end{figure}

%own device这里应该没问题，别人的论文里也有这么描述客户端之间异质性。原句：In particular, as each device generates its own local data, statistical heterogeneity is common with data being non-identically distributed between devices。

However, each client collects the local data in its own manner (\eg, different devices, different ways). Consequently, the data distribution among different clients is heterogeneous (see Fig.~\ref{fig:1} (a)), resulting in serious performance degradation of conventional algorithms~\cite{diao2020heterofl} (\eg, FedAvg~\cite{mcmahan2017communication}). Many FL algorithms have been suggested
to solve the problem of inter-client data heterogeneity for making them more robust in the non-i.i.d. setting~\cite{li2020federated,karimireddy2020scaffold,li2021fedbn,li2021model,gao2022feddc} (see Fig.~\ref{fig:1} (c)). 
% This prompts most of the existing FL algorithms to focus on how to solve the problem of data heterogeneity for making them more robust to the non-i.i.d. setting (see Fig.~\ref{fig:1} (c))~\cite{li2020federated,karimireddy2020scaffold,li2021fedbn,li2021model,gao2022feddc}. 

Actually, even for an individual client, the data may be collected by different devices~\cite{feng2021task}, under different environmental conditions, \etc. Therefore, \textit{intra-client data heterogeneity} can also be observed on complex real-world
data~\cite{chen2019blending} (see Fig.~\ref{fig:1} (b)), and give rise to serious degradation of FL performance (see Fig.~\ref{fig:1} (d)).
% In contrast, in this study, we identify a critical problem of FL: there is not only inter-client data heterogeneity but also intra-client heterogeneity {\color{red} because of the multiple mixed unknown sub-domains caused by the complex real-world environment (see Fig.~\ref{fig:1} (b))~\cite{chen2019blending}}. More importantly, intra-client data heterogeneity may greatly degrade the FL performance, as shown in Fig.~\ref{fig:1} (d). 
A naive solution is to build a specific model for each instance
% so that the local model can be personalized for different instances
~\cite{zou2022learning}. However, such a scheme results in significant challenges in the training, \textit{storage and communication} of the instance-adaptive models. Instead, we resort to the parameter-efficient fine-tuning, \eg, prompt tuning~\cite{jia2022visual}, of pre-trained models. As for prompt tuning, we can simply freeze the backbone of pretrained models, only learn and communicate a small number of learnable prompts between the server and the client~\cite{feng2023learning,shin2020autoprompt,schick2020s,jia2022visual}. Nonetheless, instance-wise modeling remains an \textit{unstudied} issue for parameter-efficient fine-tuning. Additionally, in comparison to the original pre-trained model, prompt tuning also introduces \textit{additional} parameters and \textit{increases} the computation cost in the inference stage~\cite{lian2022scaling}.

% costs, especially for large-scale pre-training models, which can be restrictive since the model parameters need to be updated on both the client and server sides. Although we can simply use the prompt tuning paradigm to freeze the large-scale pre-training model and only learn and communicate a small number of learnable parameters between the server and the client~\cite{shin2020autoprompt,schick2020s,jia2022visual}, instance-wise modeling still leaves the storage and communication cost problem for the local end~\cite{lian2022scaling}. Additionally, the prompt tuning paradigm is not only sensitive to the number of prompts for each client but also introduces additional parameters and computation cost in the inference stage compared to the original pre-trained model~\cite{lian2022scaling}.

To handle the intra-client data heterogeneity, instead of instance-adaptive models, this paper resorts to \textit{\textbf{instance-adaptive inference}}, and presents a novel FL algorithm, \ie, FedIns, built upon pre-trained models. For parameter-effective fine-tuning, following~\cite{lian2022scaling}, we scale and shift the deep features (SSF) to fine-tune the pre-trained model in the local training phase and merge them into the original pre-trained model weights by reparameterization in the inference phase. To enable instance-adaptive inference, we train a SSF pool for each client and aggregate them on the server, and thus \textit{low storage and communication costs} can still be maintained. For example, compared with FedAvg~\cite{mcmahan2017communication}, FedIns only has less than $15\%$ communication costs. Our federated SSF pool is aggregated from multiple groups of local SSF, which implicitly add client-relevant knowledge to the \textit{\textbf{federated SSF pool}} and dynamically guide the client to handle the corresponding response in an instance-wise fashion. During inference, for a given instance, we dynamically find the best-matched SSF subsets from the pool and aggregate them to generate an adaptive SSF specified for the instance. As such, the model can reduce both inter- and intra-client heterogeneity, achieving a $6.64\%$ gain against FedAvg on CIFAR-100.

\vspace{2pt}
In a nutshell, our contributions are three-fold:
\begin{itemize}[leftmargin=*]
	\setlength{\itemsep}{0pt}
	\setlength{\parsep}{-2pt}
	\setlength{\parskip}{-0pt}
	\setlength{\leftmargin}{-15pt}
	\vspace{-5pt}
\item A novel FL algorithm, \ie, FedIns, is presented to handle \textit{intra-client data heterogeneity}, which has been overlooked by the existing FL literature.

% We conduct a systematic study on data heterogeneity in FL, and find that there is not only inter-client data heterogeneity but also intra-client heterogeneity caused by the natural data variations due to varied sub-domains in the real world, which also significantly degrades FL performance.
    %{We conduct a systematic study on data heterogeneity in FL, and find that there is not only inter-client data heterogeneity but also intra-client heterogeneity caused by the different collection equipments of the same institution, which also significantly degrades FL performance.}
\item Federated SSF is proposed and extended by allowing each client to have an SSF pool, and instance-adaptive inference is fulfilled by dynamically finding and aggregating the best matched SSF subsets for each test instance.
%    {By introducing an SSF pool to each client, we aggregate these SSF pools on
% the server side. For a given instance, we dynamically find and aggregate the best-matched SSF subsets to enable \textit{instance-adaptive inference}.}
   % {We provide a feasible instance-adaptive model, \ie FedIns, for FL to address both the inter- and intra-client data heterogeneity. FedIns uses a parameter-efficient fine-tuning mechanism on the basis of a large-scale pre-training model to learn a powerful federated SSF pool in an instance-wise fashion, then dynamically finds relevant SSF subsets according to the instance input features, and finally aggregates these SSF pools on the server, thereby alleviating the local storage and communication costs.}
\item {Extensive experimental results show FL performance can be effectively improved by alleviating the intra- and inter-client data heterogeneity.}

   % {We evaluate the performance of FedIns against various state-of-the-art FL algorithms on different federated settings. The results indicate that FL performance can be effectively improved by alleviating the intra- and inter-client data heterogeneity.}

% \vspace*{-7pt}
\end{itemize}

\section{Related Work}

% \vspace{-2pt}
\noindent\textbf{Data Heterogeneity in FL.} Generally, federated learning algorithm aims to obtain an aggregated model that minimizes training losses for all clients. The classic FL algorithm, FedAvg~\cite{mcmahan2017communication} simply sends the local model to the server for aggregation to learn a global model. However, as each device generates its own local data, data heterogeneity across different clients occurs, making FedAvg becomes a sub-optimal solution for FL~\cite{li2019convergence,yuan2022addressing,lee2021preservation,feng2022specificity}. 

Existing FL methods usually alleviate this problem in two aspects. 
One is to improve local training. For example, FedProx adds a proximal term to the objective function of the local model to tackle heterogeneity~\cite{li2020federated}. 
Karimireddy \textit{et al.} used control variates to correct the client-drift caused by data heterogeneity~\cite{karimireddy2020scaffold}. Li \textit{et al.} kept all the batch normalization in the local to alleviate the heterogeneity of local data across clients~\cite{li2021fedbn}. Inspired by contrastive learning, Li \textit{et al.}~\cite{li2021model} corrected the local clients by computing similarity between model representations to handle the heterogeneity issue. Gao \textit{et al.} used an auxiliary local drift variable to bridge the gap between the local and the global model parameters, thereby alleviating the data heterogeneity~\cite{gao2022feddc}. Mendieta \textit{et al.} alleviated data heterogeneity by promoting local learning generality rather than proximal restriction~\cite{mendieta2022local}. 
The non-i.i.d. problem caused by data heterogeneity can also be alleviated by addressing catastrophic forgetting from the server to the client, where each local and global communication is regarded as a new task~\cite{xu2022acceleration,lee2022preservation}.

Besides improving local training, another is to improve the server aggregation process for alleviating data heterogeneity. 
For example, Yurochkin \textit{et al.} replaced classical aggregation schemes by matching neuron aggregation in local models based on a Bayesian non-parametric approach~\cite{yurochkin2019bayesian}. Analogously, Wang \textit{et al.} created a normalized averaging method as an alternative to the average update mechanism~\cite{wang2020tackling}. % 
To sum up, to handle the data heterogeneity across clients, existing methods either limit the impact of local updates on the server (\eg, by regularizing and personalizing the design of clients to correct the update direction of locals~\cite{li2020federated,karimireddy2020scaffold,li2021fedbn,li2021model,gao2022feddc,mendieta2022local}), or modify the aggregation scheme~\cite{yurochkin2019bayesian,wang2020tackling}.

However, in many real-world applications, local data in each client may be collected by different devices and from different environments.
Consequently, there may be \textit{multiple mixed unknown sub-domains within one client}~\cite{chen2019blending}, \ie, intra-client data heterogeneity.
%
%However, the real-world environment often includes extensive arrays of scenarios and continuously changes as time goes by, which means that there may be \textit{multiple mixed unknown sub-domain that contain intra-client cases}~\cite{chen2019blending}.
%
As discussed in Sec.~\ref{sec:intro}, both intra-client and inter-client data heterogeneity will lead to the performance degradation of FL. 
Nonetheless, existing methods mainly focus on inter-client data heterogeneity, leaving \textit{intra-client heterogeneity remain an uninvestigated issue}. 
Given this, this work presents a novel FedIns algorithm to handle intra-client data heterogeneity, which can also be readily deployed to alleviate inter-client data heterogeneity.

% \vspace{-2pt}
\noindent\textbf{Parameter-efficient Fine-tuning.} In the recent few years, we have witnessed the promising performance of pre-training models in many downstream tasks. Accordingly, in the field of FL~\cite{nguyen2022begin,chen2022pre}, pre-trained models can also serve as a strong baseline. 
However, the number of parameters of pre-trained models is usually very large~\cite{shu2021zoo,han2021pre}, and simply fine-tuning the full model undoubtedly yields a huge amount of communication cost in FL algorithms. Therefore, some works have attempted to explore large-scale pre-training models in a parameter-efficient fine-tuning manner~\cite{kim2021vilt}. In this way, the backbone network is frozen during training, and only a small number of parameters can be learned or tuned to ``understand'' downstream tasks. 
In particular, prompt~\cite{lester2021power,jia2022visual,li2022prompt}, adapter~\cite{houlsby2019parameter,he2021towards,nie2022pro}, and SSF~\cite{lian2022scaling} have been suggested to leverage the representation abilities of large-scale pre-training models to achieve good performance on downstream tasks by fine-tuning a few trainable parameters.

Among these parameter-efficient fine-tuning methods, most prompt and adapter-based methods~\cite{houlsby2019parameter,karimi2021compacter} introduce additional parameters and computational costs in the inference stage. 
In addition, prompt tuning is \textit{sensitive} to the number of prompts; \eg, different tasks need to adopt different numbers of prompts, and an inappropriate number will reduce the accuracy or increase the redundancy of the calculation. 
In contrast, SSF~\cite{lian2022scaling} does not bring extra parameters and FLOPs during the inference phase since it only adds learnable parameters during the training phase and merges them into the original pre-trained model weights via reparameterization after training. 
Despite these progress, it remains not studied for the effectiveness of SSF in a distributed framework with intra- and inter-client data heterogeneity. Therefore, different from the above centralized works, in this paper we focus on exploring how to \textit{make SSF work for decentralized FL frameworks} to effectively handle data heterogeneity.

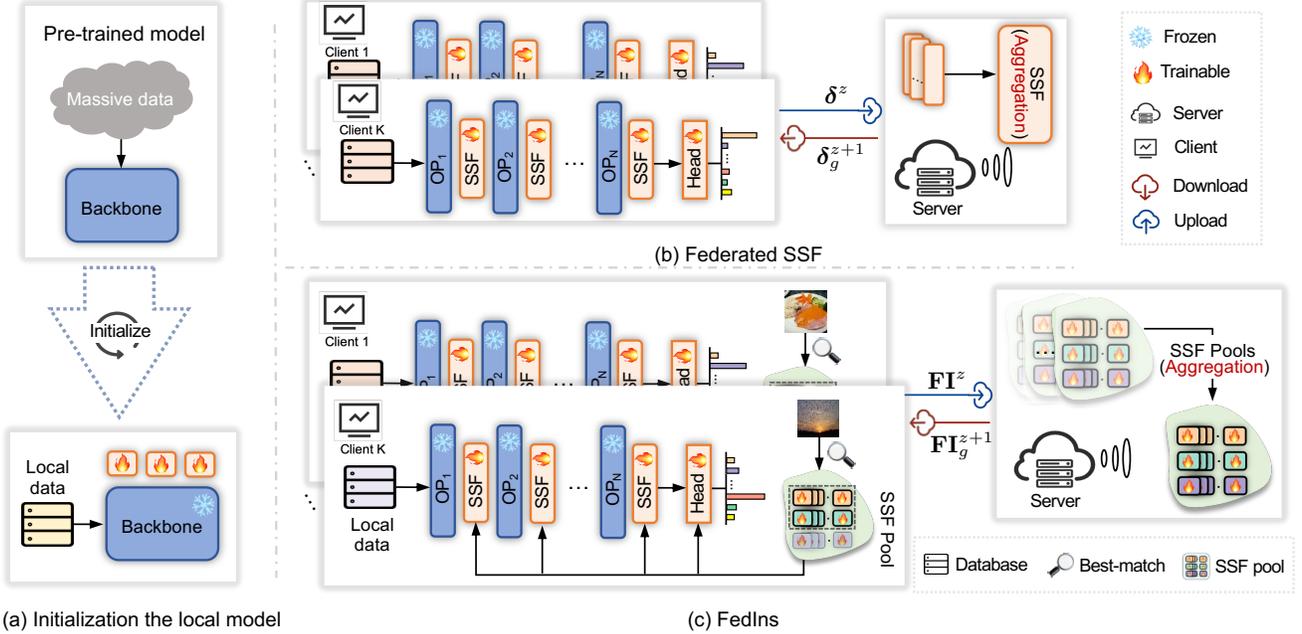
\begin{figure*}[!t]
    % \vspace{-2pt}
	\begin{center}
		\includegraphics[width=0.99\linewidth]{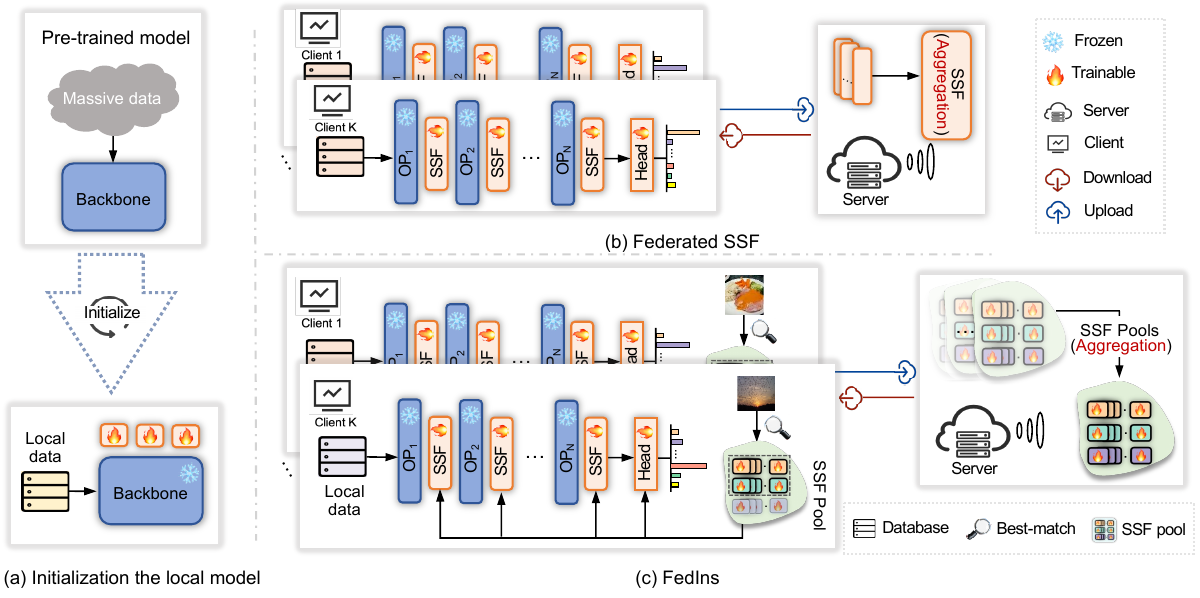} 
          \put(-182,203){ \small$\bm{\delta}^{z}$}
          \put(-185,180){ \small$\bm{\delta}_g^{z+1}$}
          \put(-143,95){ \small${\mathbf{FI}}^{z}$}
          \put(-142,71){ \small${\mathbf{FI}}_g^{z+1}$}   
	\end{center}
    \vspace{-14pt}
	\captionsetup{font=small}
	\caption{\textbf{Overall pipeline} of our proposed \textbf{FedIns}. (a) Pre-train the backbone on the massive data \textit{offline} and fine-tune a small number of learnable parameters on local data in a parameter-efficient manner. (b) For federated SSF, we train an SSF for each client and aggregate the SSFs from all clients to obtain the global SSF (see Sec.~\ref{sec:fedssf}). (c) As for FedIns, we train a federated SSF pool for \textit{each client} and aggregate the SSF pools from all clients to obtain the global SSF pool (see Sec.~\ref{sec:pool}).}
	\vspace{-11pt}
	\label{fig:2}
\end{figure*}

\section{Methodology}\label{sec:method}
% \vspace{-2pt}
% Before detailing our proposed algorithm FedPR for MRI reconstruction (\S\ref{sec:fedpr} and \S\ref{sec:null}), we first introduce the notion of federated MRI (\S\ref{sec:federatedmri}).
% \vspace{-2pt}
\subsection{Federated Learning}\label{sec:FL_statement}
% \vspace{-3pt}
As growing privacy concerns arise, federated learning has received intensive recent interest by training and deploying deep neural network models in a distributed manner~\cite{mcmahan2017communication}. Suppose there are $K$ local clients, where each of them has their local dataset $\mathcal{D}^{k}$. The distributed paradigm FL aims to learn a global model $\boldsymbol{w}$ over the whole training data $\mathcal{D} = \left\{\mathcal{D}^{1}, \mathcal{D}^{2}, \ldots, \mathcal{D}^{K}\right\}$
using a central server without exchanging local private data. Formally, such a process can be expressed as
\vskip -10pt
\begin{equation}\label{eq:1}
{\arg \underset{\boldsymbol{w}}\min } \mathcal{L}(\boldsymbol{w})= \sum_{k=1}^K \frac{|\mathcal{D}^k|} {|\mathcal{D}|} \mathcal{L}_k(\boldsymbol{w}),
% \vspace{-2pt}
\end{equation}
where $|\mathcal{D}|$ denotes the number of samples in $\mathcal{D}$, and $\boldsymbol{w}$ denotes the model parameters. $\mathcal{L}_k(\boldsymbol{w})$ is the empirical loss of client $k$ which can be expressed as
\vspace{-2pt}
\begin{equation}\label{eq:2}
\mathcal{L}_{k}(\boldsymbol{w}_{{}})=\mathbf{E}_{\left(\mathbf{x}, \mathbf{y}\right) \in \mathcal{D}^k} \ell_k\left(\mathbf{x} ;\boldsymbol{w}\right),
\vspace{-1pt}
\end{equation}
where $\ell_k$ denotes the local loss term, \eg, cross-entropy loss, and $\mathbf{x}$ denote a sample of client $k$.

% \vspace{-2pt}
\subsection{Learning Federated SSF}\label{sec:fedssf}
% \vspace{-2pt}
In federated learning, the local data $\mathcal{D}^{k}$ of a client may not be sufficient to train a large scale deep network. 
Pre-trained models can thus be introduced to compensate for the deficiency of local data~\cite{nguyen2022begin,chen2022pre}.
%
%The complexity and variety of local clients usually lead to insufficient training data $\mathcal{D}^{k}$ for some clients. However, pre-trained models have been shown to be able to close the gap between federated and centralized performance on a small amount of training data by compensating for this deficiency with their superior performance~\cite{nguyen2022begin,chen2022pre}. 
%
However, pre-trained model usually has a large number of model parameters.
Direct fine-tuning the full model consequently gives rise to significant communication costs between the server and the client. 
In this work, we adopt a recent parameter-efficient fine-tuning paradigm, \ie, SSF~\cite{lian2022scaling}, which trains only a small number of learnable parameters and brings no inference overhead by reparameterizing them into the original pre-trained model weights. 
In the following, we will introduce federated SSF, which incorporates SSF into the FL framework.

%Naturally, we can improve the robustness of FL paradigms with the help of parameter-efficient fine-tuning mechanisms.

In SSF, given a pre-trained model with parameters $\boldsymbol{\theta}$, we remodulate features by insert SSF with the scale $\bm{\gamma}$ and shift $\bm{\beta}$ factors after each operation (OP)~\cite{lian2022scaling}, \ie, multi-head self-attention (MSA), MLP and layer normalization (LN), \etc. 
During the fine-tuning phase, the model parameters of SSF can be represented as $\boldsymbol{w}\!=\!\{ \bm{\gamma},\bm{\beta},\bm{h},\boldsymbol{\theta} \}$, where $\bm{h}$ is the parameters of the classification head.
In particular, the pre-trained weights are kept frozen, and only the SSF and classification head are kept updated. 
Once the training is accomplished, $\{ \bm{\gamma},\bm{\beta},\bm{h} \}$ can then be merged into $\boldsymbol{\theta}$ to obtain the updated model parameter $\boldsymbol{\theta}^{\prime}$.

As shown in Fig.~\ref{fig:2} (b), when incorporating SSF into FL, we only require to update the client-specific SSF and classification head, \ie, $\bm{\delta} = \{ \bm{\gamma},\bm{\beta},\bm{h}\}$.
Thus, Eq.~\eqref{eq:1} can be modified as,
%
%During the local fine-tuning, the pre-trained weights are kept frozen and only the client-specific SSF and classification head are kept updated and communicated between the server and client. As a result, we can reformulate Eq.~\eqref{eq:1} as 
\vskip -8pt
\begin{equation}\label{eq:3}
\bm{\delta}_g = {\arg}~\underset{\bm{\delta}}{\min } \mathcal{L}(\bm{\delta})= \sum_{k=1}^K \frac{|\mathcal{D}^k|}{|\mathcal{D}|} \mathcal{L}_k(\bm{\delta}).
% \vspace{-2pt}
\end{equation}

Federated SSF only requires a small number of parameters $\bm{\delta}_k$ of local clients to be updated and communicated with the server, and thus is both parameter- and communication-efficient for FL.

%Such a mechanism offers both parameter- and communication-efficient and effective ways for FL because only a few learnable parameters need to be communicated between the server and the client.

\noindent{\textbf{Local Update Step:}} Assume there are $Z$ rounds of communication with $T$ local updates per round. The clients are optimized using the following update rules with a learning rate of $\eta_k$ for each communication round $z=\left\{1,2,...,Z\right\}$:
% \vskip -2pt
\begin{equation}\label{eq:5}
\boldsymbol{\bm{\delta}}_{{k}}^{z,t+1} \leftarrow \boldsymbol{\bm{\delta}}_{{k}}^{z,t}-\eta_{k} \nabla \ell_k\left(\mathbf{x}^{k};\boldsymbol{\bm{\delta}}_{{k}}^{z,t}\right),
% \vspace{-1pt}
\end{equation}
where $t$ denotes the $t$-th update of the local clients.

\noindent{\textbf{Server Update Step:}} The server performs aggregation every round by receiving the updated parameters of all participated clients after the local updates within each round. Formally, we have  
\vskip -7pt
\begin{equation}\label{eq:6}
\boldsymbol{\bm{\delta}}_{{g}}^{z+1}\leftarrow \sum_{k=1}^K\frac{|\mathcal{D}^k|}{|\mathcal{D}|} \boldsymbol{\bm{\delta}}_{{k}}^{z},
\vspace{-4pt}
\end{equation}
where $\boldsymbol{\bm{\delta}}_{{g}}^{z+1}$ denotes the global updated parameters of round $z+1$. Then, we can obtain a robust global model parameterized by $\boldsymbol{\bm{\delta}}_{{g}}$ after $Z$ rounds of communication without disclosing any local private data. 

% 这里解释re-parameterization
When the training is accomplished, we can re-parameterize the SSF by merging it into the original parameter space (\ie, model weight $\boldsymbol{\theta}$). As a result, federated SSF is not only efficient in terms of communication costs, but also does not introduce any extra parameters during the inference phase. 

% As a result, federated SSF is not only efficient in communication cost, but also bring no extra parameters in the inference phase.

\subsection{FedIns}\label{sec:pool}
% \vspace{-4pt}

In this subsection, we further present FedIns for handling intra-client data heterogeneity. 
In many complex real-world scenarios, the data in a local client may contain multiple unknown mixed sub-domains~\cite{chen2019blending}. 
Despite its merit on communication cost, federated SSF is still limited in alleviating the intra-client data heterogeneity issue.
To address this issue, we extend federated SSF to SSF pool for enabling instance-adaptive inference in the FL framework, resulting in our FedIns algorithm.
As shown in Fig.~\ref{fig:2} (c), we train an SSF pool for each client, and aggregate them into the federated SSF pool on the server.
For a given instance, we dynamically find the best-matched SSF subsets from the pool and aggregate them to generate an adaptive SSF for instance-adaptive inference.
In the following, we will introduce FedIns in more detail.

%To fill this gap, we propose a SSF pool mechanism, making instance-adaptive models feasible for FL to alleviate the intra-client data heterogeneity issue. 

%Specifically, we respectively deploy a federated SSF pool for each client, match the relevant SSF subsets dynamically for each specific input in an instance-wise manner, and finally aggregate these SSF pools only on the server. Such a process modulates the client model for each different instance, enabling the final global model to relieve both inter-client and intra-client data heterogeneity.

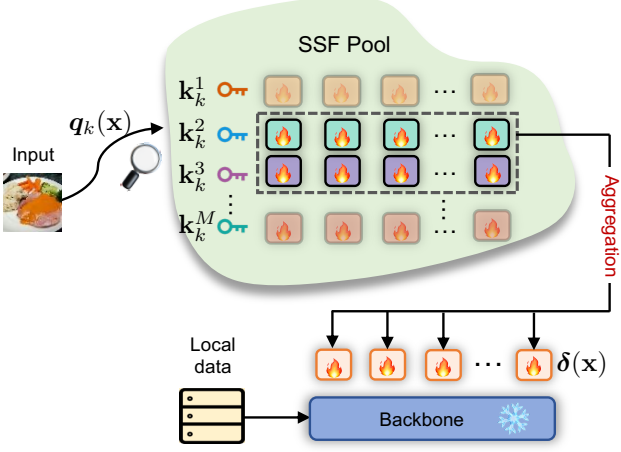
\begin{figure}[t]
    % \vspace{-8pt}
	\begin{center}
		\includegraphics[width=\linewidth]{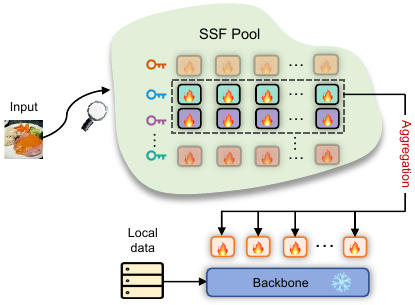}
            \put(-210,122){$\boldsymbol{\bm{q}}_k(\mathbf{x})$}
            \put(-171,134){ $\mathbf{k}_k^1$}
            \put(-171,118){ $\mathbf{k}_k^2$}
		\put(-171,102){ $\mathbf{k}_k^3$} 
            \put(-171,83){ $\mathbf{k}_k^M$}
            \put(-28,31){ $\boldsymbol{\bm{\delta}}(\mathbf{x})$}     
	\end{center}
	\vspace{-14pt}
	\captionsetup{font=small}
	\caption{\textbf{Illustration} of instance-adaptive inference. For a given instance,  we use the pre-trained model to generate a query, and then dynamically select the $C$ best-matched SSFs, which are then aggregated to form an instance-adaptive SSF for enabling instance-adaptive inference.}
	\vspace{-14pt}
	\label{fig:3}
\end{figure}

To extend federated SSF, we allow each client have an SSF pool $\boldsymbol{\bm{\Delta}}_k=\left\{\boldsymbol{\bm{\delta}}_k^1,\boldsymbol{\bm{\delta}}_k^2, \cdots, \boldsymbol{\bm{\delta}}_k^M\right\}$ of a set of $\boldsymbol{\bm{\delta}}_k^m$s, where $M$ denotes the size of pool.
Furthermore, we introduce a pool of learnable \texttt{Key}s $\mathbf{K}_k=\left\{\mathbf{k}_k^1,\mathbf{k}_k^2, \cdots, \mathbf{k}_k^M
\right\}$ corresponding to $\boldsymbol{\bm{\Delta}}_k$.  
Taking both $\boldsymbol{\bm{\Delta}}_k$ and $\mathbf{K}_k$ into account, the learnable parameters of FedIns of each client can be expressed as,
\vspace{-2pt}
\begin{equation}\label{eq:7}
\mathbf{FI}_k=\left\{\mathbf{K}_k, \boldsymbol{\bm{\Delta}}_k\right\}.
% \vspace{-4pt}
\end{equation}
Then, the local update in FedIns can be written as,
\vskip -2pt
\begin{equation}\label{eq:8}
\mathbf{FI}_{{k}}^{z,t+1} \leftarrow \mathbf{FI}_{{k}}^{z,t}-\eta_{k} \nabla \ell_k\left(\mathbf{x}^{k};\mathbf{FI}_{{k}}^{z,t}\right).
\vspace{-2pt}
\end{equation}
After $T$ local updates, the server update can be expressed as,
\vskip -13pt
\begin{equation}\label{eq:9}
{\mathbf{FI}}_g^{z+1}\leftarrow \sum_{k=1}^K\frac{|\mathcal{D}^k|}{|\mathcal{D}|} {\mathbf{FI}}_k^{z}.
\vspace{-1pt}
\end{equation}
After $Z$ rounds of communication, we can obtain the global model of FedIns, which is also parameterized by an SSF pool $\boldsymbol{\bm{\Delta}}_g$ and a set of keys $\mathbf{K}_g$. 
The detailed FedIns algorithm is given in Algorithm~\ref{alg:fedins}.
In contrast to federated SSF, during the inference phase, FedIns allow to dynamically find the best matched SSF subsets from the SSF
pool for a given instance, thereby making instance-adaptive inference feasible.

\begin{algorithm}[!t]

    \caption{FedIns}
    \label{alg:fedins}
    \KwIn{Local datasets of $K$ clients: $\mathcal{D}^{1}, \mathcal{D}^{2}, \ldots, \mathcal{D}^{K}$, local updates $T$, communication rounds $Z$, pre-trained model parameters $\boldsymbol{\theta}$, learnable parameters ${\mathbf{FI}_k}=\left\{\mathbf{K}_k, \boldsymbol{\bm{\Delta}}_k\right\}$, learning rate $\eta_{}$, hyperparameter $c$, and $M$;}
    \KwOut{The final global model $\boldsymbol{w}_{}$;}
% \KwOut{The final global model $\boldsymbol{w}_{G}$}
// \textbf{ServerExecution:}\\
Initialize global SSF pool $\boldsymbol{\bm{\Delta}}_k=\left\{\boldsymbol{\bm{\delta}}_k^1,\boldsymbol{\bm{\delta}}_k^2, \cdots, \boldsymbol{\bm{\delta}}_k^M\right\}$ with \texttt{Key}s of $\mathbf{K}_k=\left\{\mathbf{k}_k^1,\mathbf{k}_k^2, \cdots, \mathbf{k}_k^M\right\}$;
% with pre-trained model $\boldsymbol{\theta}$

\For {each communication round $z\in\left\{1,2,...,Z\right\}$}{
    \For{each client $k\in\left\{1,2,...K\right\}$ in parallel}{
             ${\mathbf{FI}}_k^{z}\leftarrow{\mathbf{FI}}_g^{z}$;\\
             Local updating with regard to each input instance $\mathbf{x}$:      ${\mathbf{FI}}_g^{z,t+1}\leftarrow \text{LocalUpdate}(k,\mathbf{x},{\mathbf{FI}}_g^{z,t})$;
    }
    
    % $\boldsymbol{\bm{\delta}}_{{g}}^{z+1}\leftarrow \sum_{k=1}^K\frac{|\mathcal{D}^k|}{|\mathcal{D}|} \boldsymbol{\bm{\delta}}_{{k}}^{z}$;\\

    ${\mathbf{FI}}_g^{z+1}\leftarrow \sum_{k=1}^K\frac{|\mathcal{D}^k|}{|\mathcal{D}|} {\mathbf{FI}}_k^{z}$;\\
   %  $\overline{\mathbf{SSF}}^{\texttt{Key}}_g\leftarrow \sum_{k=1}^K\frac{|\mathcal{D}^k|}{|\mathcal{D}|} \overline{\mathbf{SSF}}^{\texttt{Key}}_k
   % $;\\
    % Compute the uncentered covariance matrix:\\
    % $\boldsymbol{\Sigma}_{g}^{z+1}=\left(\boldsymbol{\bm{\gamma},\bm{\beta}}_{g}^{z+1}\right)\!^{\top} \boldsymbol{\bm{\gamma},\bm{\beta}}_{g}^{z+1}$;\\
    %  Approximate the null space of $\boldsymbol{\Sigma}_{g}^{z+1}$:\\
    %  $\boldsymbol{\Sigma}_{g}^{z+1} = \bm{U}^{z+1} \boldsymbol{\Lambda}^{z+1} \left(\bm{U}^{z+1}\right)\!^{\top}$;\\
    %  Select the smallest diagonal singular values of $\bm{\Lambda}_2^{z+1}$ with the ratio of $\gamma$;\\
    %  Get $\bm{U}_2^{z+1}$ corresponding to $\bm{\Lambda}_2^{z+1}$.
}
% return $\overline{\mathbf{SSF}}^{\texttt{Key}}_g$ 
return ${\mathbf{FI}}_g^{z+1}$

// \textbf{LocalUpdate} ($k$, $\mathbf{x}$, $\mathbf{FI}_{{k}}^{z,t}$)\textbf{:}

    \For{each local epoch $t\in\left\{1,2,...T\right\}$}
            {
     %                  Get the updated parameters $\Delta \boldsymbol{\bm{\gamma},\bm{\beta}}_{k}^z$:\\
     % $\Delta \boldsymbol{\bm{\gamma},\bm{\beta}}_{k}^z= \bm{U}_2^{z+1}\left(\bm{U}_2^{z+1}\right)\!^{\top} \boldsymbol{\bm{\gamma},\bm{\beta}}_{{k}}^{z}$;\\
                 $\mathbf{FI}_{{k}}^{z,t+1} \leftarrow \mathbf{FI}_{{k}}^{z,t}-\eta_{k} \nabla \ell_k\left(\mathbf{x}^{k};\mathbf{FI}_{{k}}^{z,t}\right)$\;
                 
                  % $\widetilde{\boldsymbol{\bm{\delta}}}^{z,t+1}_{k,\mathbf{x}_i}  \leftarrow \widetilde{\boldsymbol{\bm{\delta}}}_{{k,\mathbf{x}_i}}^{z,t}-\eta_{k} \nabla \ell_k\left(\mathbf{x}_i^{k};\widetilde{\boldsymbol{\bm{\delta}}}_{{k,\mathbf{x}_i}}^{z,t}\right)$\;        
% \begin{shaded}
                  %$\mathcal{L}_{k}  \leftarrow (1-\alpha)\ell_k$ 
                  %\\ $+$ $\alpha(1-\sum_{i=1}^c \operatorname{CoSim}\left(q(\mathbf{x}), \mathbf{k}^{i}\right))$ {\textcolor{magenta}{Using Eq.\eqref{eq:0}}};
% \end{shaded}
}
return 
$\mathbf{FI}_{{k}}^{z,t+1}$
% $\widetilde{\boldsymbol{\bm{\delta}}}^{z,t+1}_{k,\mathbf{x}_i}$
\end{algorithm}

\subsection{Instance-Adaptive Inference}\label{sec:inference}
% \vspace{-4pt}

When the training of FedIns is accomplished, the learned federated SSF pool $\boldsymbol{\bm{\Delta}}_g$ and the corresponding global \texttt{Key}s $\mathbf{K}_g$ will be distributed to each client.
As shown in Fig.~\ref{fig:3}, to enable instance-adaptive inference using FedIns, we resort to generating an instance-adaptive SSF from $\boldsymbol{\bm{\Delta}}_g$ and $\mathbf{K}_g$ during the inference phase.
%
%
%we define an SSF pool, $\boldsymbol{\bm{\Delta}}_k=\left\{\boldsymbol{\bm{\delta}}_k^1,\boldsymbol{\bm{\delta}}_k^2, \cdots, \boldsymbol{\bm{\delta}}_k^M\right\}$, for each client, where $M$ is the federated SSF pool size. 
%nian:\delta按前文定义也是包含了classification head的参数h的，但是这里定义的是SSF pool？确定每个pool都包含一个单独的classification head？
%Then, we define a \texttt{Key}: $\mathbf{K}=\left\{\mathbf{k}^1,\mathbf{k}^2, \cdots, \mathbf{k}^M \right\}$, for each SSF to determine the relationship that best matches the components of the SSF pool with the input instance. Formally, the SSF pool of each client associated with $\mathbf{K}$ can be expressed as
%\begin{equation}\label{eq:7}
%\mathbf{SSF}^\texttt{Key}_k=\left\{\left(\mathbf{k}, \boldsymbol{\bm{\delta}}\right)_k^1,\left(\mathbf{k}, \boldsymbol{\bm{\delta}}\right)_k^2, \cdots, \left(\mathbf{k}, \boldsymbol{\bm{\delta}}\right)_k^M\right\}.
%\vspace{-4pt}
%\end{equation}
%nian: \left(\mathbf{k}, \boldsymbol{\bm{\delta}}\right)代表什么意思？
%
%Our primal goal is to push each instance adaptively select the best-matched SSF elements from $\mathbf{SSF}^\texttt{Key}_k$. 

To this end, for a given instance $\mathbf{x}$ of a client, we first use the pre-trained model to generate a query $\boldsymbol{\bm{q}}_k(\mathbf{x})$ which has the same dimension as the keys. 
Based on the cosine similarity between $\boldsymbol{\bm{q}}_k(\mathbf{x})$ and each keys in $\mathbf{K}_g$, we select the $C$ best-matched SSFs, \ie, $\left\{\boldsymbol{\bm{\delta}}_g^{m_1},\boldsymbol{\bm{\delta}}_g^{m_2}, \cdots, \boldsymbol{\bm{\delta}}_g^{m_C}\right\}$ from $\boldsymbol{\bm{\Delta}}_g$ corresponding to the $C$ most similar keys.  
The instance-adaptive SSF of $\mathbf{x}$ can then be given by,
\vskip -5pt
\begin{equation}\label{eq:10}
\boldsymbol{\bm{\delta}}(\mathbf{x}) = \sum_{c=1}^C\frac{1}{C} \boldsymbol{\bm{\delta}}_g^{m_c}.
% \vspace{-2pt}
\end{equation}
We note that $\boldsymbol{\bm{\delta}}(\mathbf{x})$ is determined by $\mathbf{x}$ and is instance-adaptive. 
Thus, instance-adaptive inference can be fulfilled by merging $\boldsymbol{\bm{\delta}}(\mathbf{x})$ into $\boldsymbol{\theta}$ to obtain the instance-adaptive model parameter $\boldsymbol{\theta}^{\prime}(\mathbf{x})$.

\section{Experiments}
% \vspace{-4pt}
\subsection{Experimental Setup}
% \vspace{-4pt}
\noindent{\textbf{Implementation Details.}} We implement our method with Pytorch on one NVIDIA RTX 3090Ti GPU. During the federated training, all participants adopt the same hyperparameter settings, \eg, $M = 25$, $C = 3$. Both collaborative and local updating use the stochastic gradient descent (SGD) optimizer with a batch size of $32$ and a learning rate of $0.01$.

\begin{table*}[t]
\renewcommand{\arraystretch}{1.3}
	\caption{\small\textbf{Accuracy} $\%$ of state-of-the-art FL methods on 
  two scenarios, including \texttt{Label Shift}: \textbf{CIFAR-100}~\cite{krizhevsky2009learning} and \textbf{Tiny-ImageNet}, and \texttt{Feature Shift}: \textbf{DomainNet}~\cite{peng2019moment}, where \textbf{\# Com.cost} is the communication cost. \textit{w.}\texttt{Full} indicates that the local model of a FL algorithm is fully fine-tuned. The arrow ${\color{ForestGreen}\uparrow}$ and ${\color{red}\downarrow}$ indicate improvements and decrements compared with FedAvg (\textit{w.}\texttt{Full}), respectively. Detailed analyses are provided in Sec.~\ref{sec:acc}.}
	\vspace{-6pt}
	\label{tab:1}
        
	\fontsize{9}{9}\selectfont
	\centering
	\begin{tabular}{l c c cc cc cc}
\toprule

\textbf{Method}
&\multicolumn{1}{c}{\textbf{~\# Com.cost~}}
&\multicolumn{2}{c}{\textbf{~~~~~DomainNet~~~~~}} 
&\multicolumn{2}{c}{\textbf{~~~~CIFAR-100~~~~}}
&\multicolumn{2}{c}{\textbf{~~~~Tiny-ImageNet~~~~}} \\

\cmidrule(r){1-1} 
\cmidrule{2-2} 
\cmidrule(){3-8} 

SOLO 
&\multicolumn{1}{|c|}{$-$ }
&\multicolumn{2}{c}{62.18\stdvd{$17.84$}} 
&\multicolumn{2}{c}{33.81\stdvd{$47.41$}}
&\multicolumn{2}{c}{17.17\stdvd{$61.85$}}\\\hline

% Centralized (Ub) &\multicolumn{1}{|c}{$-$ }&\multicolumn{1}{|c|}{$85.80$ M}  &\multicolumn{2}{c}{45.93\stdvd{$34.09$}} &\multicolumn{2}{c}{18.24\stdvd{$62.98$}}&\multicolumn{2}{c}{17.17\stdvd{$61.85$}}\\\hline

FedAvg${\color{magenta}_{2017}}$ (\textit{w.}\texttt{Full})~\cite{mcmahan2017communication} 
&\multicolumn{1}{|c|}{$85.80$ M}
% &\multicolumn{1}{|c|}{$85.80$ M}  
&\multicolumn{2}{c}{$80.02$\stdvw{$0.00$}} 
&\multicolumn{2}{c}{$81.22$\stdvw{$0.00$}}
&\multicolumn{2}{c}{$79.02$\stdvw{$0.00$}}\\

FedProx${\color{magenta}_{2020}}$~\cite{li2020federated} 
&\multicolumn{1}{|c|}{$85.80$ M}
% &\multicolumn{1}{|c|}{$85.80$ M}  
&\multicolumn{2}{c}{$78.73$\stdvd{$1.29$}} 
&\multicolumn{2}{c}{$81.56$\stdvu{$0.34$}}
&\multicolumn{2}{c}{$79.57$\stdvu{$0.55$}}\\

SCAFFOLD${\color{magenta}_{2020}}$~\cite{karimireddy2020scaffold} 
&\multicolumn{1}{|c|}{$85.80$ M}
% &\multicolumn{1}{|c|}{$85.80$ M} 
&\multicolumn{2}{c}{$80.31$\stdvu{$0.29$}} 
&\multicolumn{2}{c}{$78.02$\stdvd{$3.20$}}
&\multicolumn{2}{c}{$76.76$\stdvd{$2.26$}}\\

FedBN${\color{magenta}_{2021}}$~\cite{li2021fedbn} 
&\multicolumn{1}{|c|}{$85.76$ M}
% &\multicolumn{1}{|c|}{$85.80$ M} 
&\multicolumn{2}{c}{$80.07$\stdvu{$0.05$}} 
&\multicolumn{2}{c}{$81.24$\stdvu{$0.02$}}
&\multicolumn{2}{c}{$79.82$\stdvu{$0.80$}}\\

MOON${\color{magenta}_{2021}}$~\cite{li2021model} 
&\multicolumn{1}{|c|}{$85.80$ M}
% &\multicolumn{1}{|c|}{$85.80$ M} 
&\multicolumn{2}{c}{$81.65$\stdvu{$1.64$}} 
&\multicolumn{2}{c}{$81.92$\stdvu{$0.70$}}
&\multicolumn{2}{c}{$81.38$\stdvu{$2.18$}}\\

FedDC${\color{magenta}_{2022}}$~\cite{gao2022feddc} 
&\multicolumn{1}{|c|}{$85.80$ M}
% &\multicolumn{1}{|c|}{$85.80$ M} 
&\multicolumn{2}{c}{$79.92$\stdvd{$0.10$}} 
&\multicolumn{2}{c}{$78.44$\stdvd{$2.78$}}
&\multicolumn{2}{c}{$79.81$\stdvu{$0.79$}}\\

{\cellcolor{mypink}\textbf{FedIns} (\textit{w.}~SSF Pool) } 
&\multicolumn{1}{|c|}{{\cellcolor{mypink}$\textbf{5.35}$ M}}
% &\multicolumn{1}{|c|}{{\cellcolor{mypink}$\textbf{5.35}$ M}} 
&\multicolumn{2}{c}{{\cellcolor{mypink}$\textbf{82.34}$\stdvu{$\underline{\textbf{2.32}}$}}} &\multicolumn{2}{c}{{\cellcolor{mypink}$\textbf{84.11}$\stdvu{$\underline{\textbf{2.89}}$}}}
&\multicolumn{2}{c}{{\cellcolor{mypink}$\textbf{86.29}$\stdvu{$\underline{\textbf{7.27}}$}}}\\

{\cellcolor{mypink}\textbf{FedIns} (${\texttt{Ours}}$) } 
&\multicolumn{1}{|c|}{{\cellcolor{mypink}$\textbf{5.35}$ M}}
% &\multicolumn{1}{|c|}{{\cellcolor{mypink}$\textbf{5.35}$ M}} 
&\multicolumn{2}{c}{{\cellcolor{mypink}$\textbf{83.12}$\stdvu{$\underline{\textbf{3.10}}$}}} &\multicolumn{2}{c}{{\cellcolor{mypink}$\textbf{84.83}$\stdvu{$\underline{\textbf{3.61}}$}}}
&\multicolumn{2}{c}{{\cellcolor{mypink}$\textbf{86.79}$\stdvu{$\underline{\textbf{7.77}}$}}}\\

\bottomrule
\end{tabular}
\vspace{-10pt}
\end{table*}

\vspace{3pt}
\noindent{\textbf{Datasets.}}~Experiments are conducted on two scenarios~\cite{li2021fedbn,li2022federated}, including \texttt{Label Shift}: \textbf{CIFAR-100}~\cite{krizhevsky2009learning}, and \textbf{Tiny-ImageNet}\footnote{\url{https://www.kaggle.com/c/tiny-imagenet}.}, and \texttt{Feature Shift}: \textbf{DomainNet}~\cite{peng2019moment}. To simulate the FL scenario, we use Dirichlet distribution to split the training data of CIFAR-100 and Tiny-ImageNet into multiple non-i.i.d. clients, respectively, and evaluate the performance on the test data~\cite{li2021model}. DomainNet consists of data from six different domains with the same class labels. Following~\cite{li2021fedbn}, we deploy the data from each domain as a specific client. To demonstrate the effectiveness of our proposed approach, each client in our experiments has only a small number of images, \ie, \textbf{DomainNet} is divided into six clients with only $26$ images for each of them; \textbf{CIFAR-100} is divided into five clients with $123$, $209$, $168$, $222$, and $278$ images, respectively; and \textbf{Tiny-ImageNet} is divided into five clients with $322$, $360$, $343$, $466$, and $509$ images, respectively.

\vspace{3pt}
\noindent{\textbf{Baselines.}}~We compare our method, FedIns, with various state-of-the-art FL algorithms, including: \textbf{(1)} FedProx~\cite{li2020federated}, which alleviates the data heterogeneity by applying a proximal term to the local objective function; \textbf{(2)} SCAFFOLD~\cite{karimireddy2020scaffold}, which corrects the client-drift by a series of control variates; \textbf{(3)} MOON~\cite{li2021model}, which corrects the local update by computing the similarity between model representations; \textbf{(4)} FedBN~\cite{li2021fedbn}, which alleviates the client-shift by using batch normalization on each local client; \textbf{(5)} FedDC~\cite{gao2022feddc}, which bridges the gap between the local and global model parameter by an auxiliary local drift variable and the classical FL algorithm; and \textbf{(6)} FedAvg~\cite{mcmahan2017communication}, which trains a global model by averaging parameters from all the participating clients. In comparison, we also add \textbf{(7)} SOLO, where participants train a model on each client and their private data without FL.
% and the (8) Centralized, where a single model trains with the combination of all the local data, into our baselines. 
% It is worth noting that the centralized server can be regarded as the upper bound of FL algorithms. 
For a fair comparison, we use ViT-B/16~\cite{dosovitskiy2020image} pre-trained on ImageNet-21K as the backbone of all the methods in our experiments.

\vspace{-4pt}
\subsection{Comparison with State-of-the-arts}\label{sec:acc}
To assess the effectiveness of our method, FedIns, we compared it to the above state-of-the-art FL algorithms, which aim to alleviate the data heterogeneity. For a fair comparison, all methods are retrained using the same model with their best hyperparameters. Specifically, for MOON, the hyperparameter of $\mu$ is set to $1$. For FedProx~\cite{li2020federated}, the hyperparameter to control the weight of its proximal term is set to $0.001$. For FedDC, we set the hyperparameter of $\alpha$ to $0.01$. Here, all the competing methods are trained with $300$ communication rounds and $10$ local epochs for each round. As the results shown in Table~\ref{tab:1}, our FedIns consistently outperform baselines in all settings, achieving state-of-the-art results in most cases. In particular, on \textbf{DomainNet}, \textbf{CIFAR-100}, and \textbf{Tiny-ImageNet} datasets, our method improves the accuracy of FedAvg from $80.02\%$, $81.22\%$, and $79.02\%$ to $\textbf{83.12}\%$, $\textbf{84.83}\%$, and $\textbf{86.79}\%$, respectively. These results indicate the robustness of our method in the two scenarios, \ie, \texttt{Label Shift} and \texttt{Feature Shift}. The FedAvg (\textit{w.}\texttt{Full})~\cite{mcmahan2017communication} is prone to overfitting, even it uses the pre-trained model locally when local data is limited. However, the parameter-efficient method of freezing the backbone network is consistent with the mechanism of personalized FL which is also an effective way to improve FL performance. More importantly, the number of training parameters and communication cost of our method are only $15\%$ of those of FedAvg (\textit{w.}\texttt{Full})~\cite{mcmahan2017communication}. It is worth noting that FedProx~\cite{li2020federated} and SCAFFOLD~\cite{karimireddy2020scaffold} perform worse than FedAvg~\cite{mcmahan2017communication}, \ie, 
DomainNet: $80.02$$\%$ $\rightarrow$ $\textbf{78.73}$$\%$ and CIFAR-100: $81.22$$\%$ $\rightarrow$ $\textbf{78.02}$$\%$. This is primarily due to the fact that correcting local updates through proximal terms and control variates will result in some deviations when there are only a few training datasets on the local clients. Differently, FedDC dynamically bridges the gap between the local model and the global model in the parameter aggregation stage, making the global model able to alleviate the data heterogeneity~\cite{gao2022feddc}. Nonetheless, even compared with FedDC~\cite{gao2022feddc}, which is the latest work aiming to address the statistical heterogeneity, our method still achieves a $8.75\%$ improvement on \textbf{Tiny-ImageNet}, (\ie, $79.81$$\%$ $\rightarrow$ $\textbf{86.79}$$\%$).

% In addition, the accuracy of FedIns is almost on par with that of Centralized, which is the upper bound of the FL algorithms, \ie, {\color{blue}$xx$ \textit{vs.}~$\textbf{xx}$}.

In contrast, the results of SOLO are lower than those of other FL algorithms, although it is also fine-tuned on a pre-trained model, which demonstrates the benefits of FL. These classical FL algorithms only consider the inter-client data heterogeneity, ignoring the intra-client data heterogeneity, which will also degrade the FL performance. This finding confirms our core idea that the performance of FL algorithms can be improved by simultaneously relieving the inter- and intra-client data heterogeneity. In the next section, we analyze the communication efficiency of FedIns and the influence of FedIns on inter-/intra-client data heterogeneity and local epoch.

\begin{figure*}[!t]
    \vspace{-2pt}
	\begin{center}
		\includegraphics[width=\linewidth]{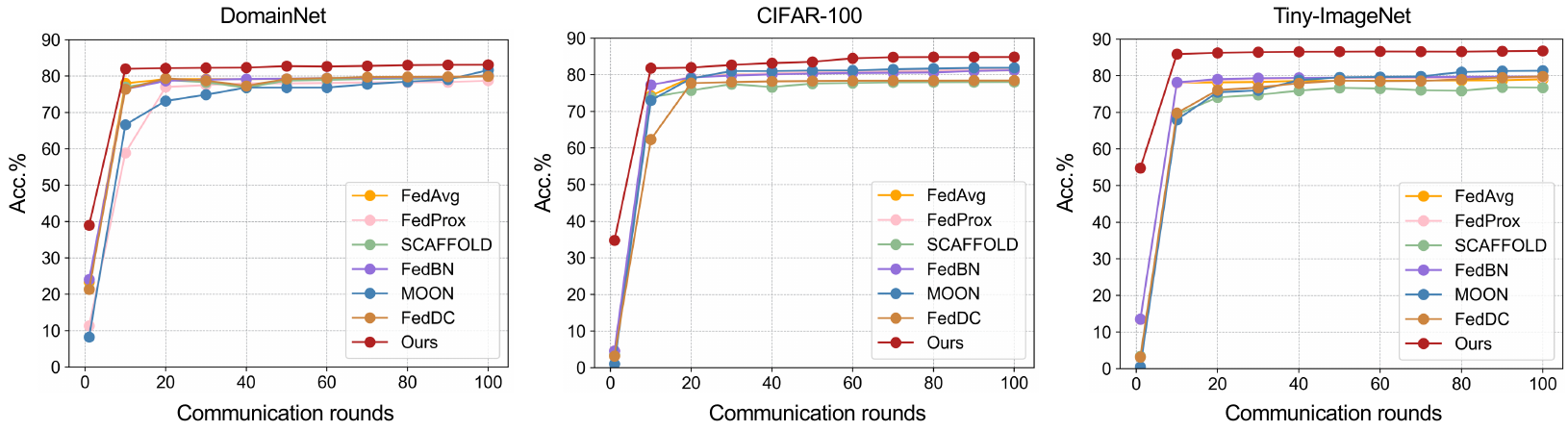}
	\end{center}
    \vspace{-18pt}
	\captionsetup{font=small}
	\caption{\textbf{Accuracy comparison} of different state-of-the-art FL algorithms in terms of different communication rounds on two scenarios, including \texttt{Label Shift}: \textbf{CIFAR-100}~\cite{krizhevsky2009learning} and \textbf{Tiny-ImageNet}, and \texttt{Feature Shift}: \textbf{DomainNet}~\cite{peng2019moment} datasets.}
	\vspace{-5pt}
	\label{fig:4}
\end{figure*}

\begin{table*}[t]
\renewcommand{\arraystretch}{1.3}
	\caption{\small\textbf{Ablation} studies with regard to the \textit{key components} of FedIns on the three datasets, where \textbf{\texttt{\#New.Param.}} represents the new introduced parameters of each method, ${\color{red}\downarrow}$ indicates decrements compared with our full model \emph{SSF Pool} (\texttt{Ours}). Detailed analyses are provided in Sec.\ref{keycom}.}
	\vspace{-8pt}
	\label{tab:2}

	\fontsize{9}{9}\selectfont
	\centering
	\begin{tabular}{l c c  c c cc cc cc cc}
\toprule
% \multicolumn{1}{l}{}&&\multicolumn{2}{c}{\textbf{\texttt{In-Federation}}}&\multicolumn{2}{c}{\textbf{\texttt{Out-of-Federation}}} \$\%$\midrule
% \cmidrule(lr){3-8} \cmidrule(l){9-14}
\textbf{Variation}&\multicolumn{1}{c}{\textbf{\texttt{Prompt}}}&\multicolumn{1}{c}{\textbf{\texttt{SSF}}}&\multicolumn{1}{c}{\textbf{\texttt{Pool}}}&\multicolumn{1}{c}{\textbf{\texttt{\#New.Param.}}}&\multicolumn{2}{c}{\textbf{~DomainNet~}} &\multicolumn{2}{c}{\textbf{~CIFAR-100~}}&\multicolumn{2}{c}{\textbf{~Tiny-ImageNet~}} \\
 \cmidrule(lr){1-1} \cmidrule(l){2-2} \cmidrule(l){3-3} \cmidrule(l){4-4} \cmidrule(l){5-5} \cmidrule(l){6-11} 
$\emph{Prompt}$  
&\multicolumn{1}{|c}{\Checkmark} 
&$-$
&$-$
&\multicolumn{1}{|c}{$0.10$ M}

&\multicolumn{2}{|c}{$79.03$\stdvd{$4.09$}}
&\multicolumn{2}{c}{$80.25$\stdvd{$4.58$}}
&\multicolumn{2}{c}{$81.02$\stdvd{$5.77$}}\\

$\emph{Prompt Pool}$ 
&\multicolumn{1}{|c}{\Checkmark}
&$-$
&\multicolumn{1}{c}{\Checkmark}
&\multicolumn{1}{|c}{$2.32$ M}

&\multicolumn{2}{|c}{$80.17$\stdvd{$2.95$}}
&\multicolumn{2}{c}{$81.39$\stdvd{$3.44$}}
&\multicolumn{2}{c}{$81.98$\stdvd{$4.81$}}\\

{\cellcolor{mypink}\emph{SSF} }
&\multicolumn{1}{|c}{{\cellcolor{mypink}$-$}}
&\multicolumn{1}{c}{{\cellcolor{mypink}\Checkmark}}
&{\cellcolor{mypink}$-$}
&\multicolumn{1}{|c}{{\cellcolor{mypink}$0.00$ M}}
&\multicolumn{2}{|c}{{\cellcolor{mypink}\textbf{$80.75$}\stdvd{${{2.37}}$}}}
&\multicolumn{2}{c}{{\cellcolor{mypink}\textbf{$83.81$}\stdvd{${{1.02}}$}}}
&\multicolumn{2}{c}{\cellcolor{mypink}{\textbf{$84.32$}\stdvd{${{2.47}}$}}}\\

{\cellcolor{mypink}\emph{SSF Pool} (\texttt{Ours}) }
&\multicolumn{1}{|c}{{\cellcolor{mypink}$-$}}
&\multicolumn{1}{c}{{\cellcolor{mypink}\Checkmark}}
&\multicolumn{1}{c}{{\cellcolor{mypink}\Checkmark}}
&\multicolumn{1}{|c}{{\cellcolor{mypink}$5.16$ M}}
&\multicolumn{2}{|c}{{\cellcolor{mypink}\textbf{$83.12$}\stdvno{${{0.00}}$}}} 
&\multicolumn{2}{c}{{\cellcolor{mypink}\textbf{$84.83$}\stdvno{${{0.00}}$}}}
&\multicolumn{2}{c}{{\cellcolor{mypink}\textbf{$86.79$}\stdvno{${{0.00}}$}}}\\
\bottomrule
\end{tabular}
\vspace{-12pt}
\end{table*}

\subsection{Communication Efficiency Analysis}
Our primary goal is to investigate how to efficiently alleviate inter- and intra-client heterogeneity and make instance-adaptive models feasible for FL, leveraging FL to perform well with less local data, a smaller number of parameters, and lower communication costs. Therefore, we evaluate the communication efficiency of the proposed method in terms of communication cost, number of parameters, and different numbers of communication rounds. As shown in Table~\ref{tab:1}, FedIns only requires $5.35$ M of communication, which is $15\%$ of the others. Similarly, the number of learnable parameters of FedIns is only $5.35$ M. Therefore, our method has high efficiency in both local client updates and server communication. On the contrary, although the classical federated algorithms MOON~\cite{li2021model} and FedBNs~\cite{li2021fedbn} can also achieve good accuracy, they require a large amount of local computation and incur high communication costs. In Fig.~\ref{fig:4}, we compare the accuracy of all methods under different communication rounds to demonstrate the superior communication efficiency of our method. We note that for all competing methods, the local epoch of each method is fixed at $10$. As can be seen from this figure, FedIns reached stability in the $10$-th round, while the other methods needed to reach stability after $20$ rounds. This indicates that our FedIns dynamically guides the local model to alleviate the intra-client data heterogeneity, enabling the instance-adaptive model feasible in FL.

\subsection{Effect of Statistical Heterogeneity}
\begin{figure}[!t]
    % \vspace{-15pt}
	\begin{center}
		\includegraphics[width=\linewidth]{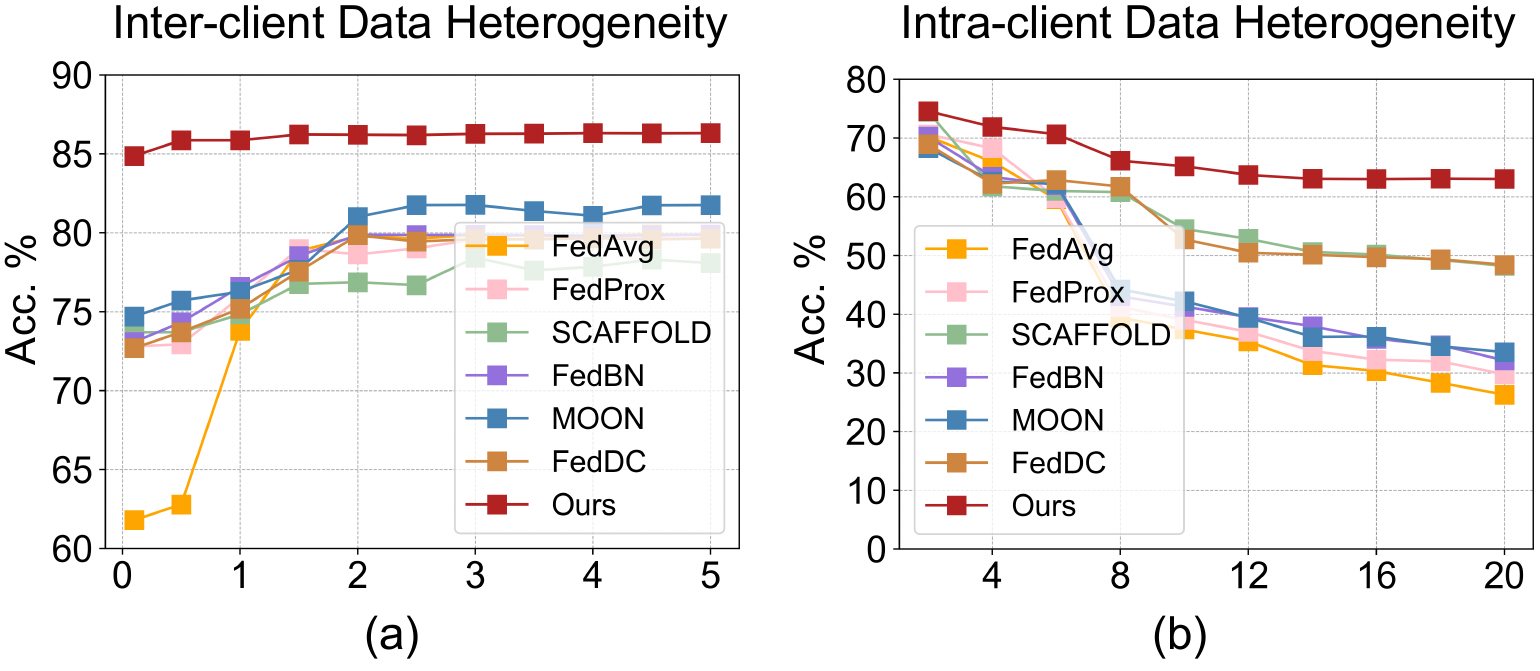}
         \put(-172,1.8){ \small$\beta$}
         \put(-50,3.7){ \small$\gamma$}
	\end{center}
    \vspace{-15pt}
	\captionsetup{font=small}
	\caption{\small\textbf{Accuracy comparison} of different state-of-the-art FL algorithms in terms of various \textbf{(a)} \textit{inter-client heterogeneity} and \textbf{(b)} \textit{intra-client heterogeneity} on \textbf{Tiny-ImageNet}, where the smaller $\beta$ values, the higher inter-client heterogeneity; the larger $\gamma$ values, the higher intra-client heterogeneity.}
	\vspace{-13pt}
	\label{fig:5}
\end{figure}

% (inter intra)
As we mentioned before, our proposed method can improve the performance of FL algorithms by alleviating both inter- and intra-client data heterogeneity. Therefore, here we explore whether the proposed method remains effective as inter- and intra-client data heterogeneity increases. First, we examine the influence of inter-client data heterogeneity on our algorithm by changing the concentration parameter $\beta$ of the Dirichlet distribution. The smaller $\beta$ values, the higher inter-client data heterogeneity~\cite{li2021model}. Fig.~\ref{fig:5} (a) shows the classification accuracies of $\beta$ values vary in $\{0.1, 0.5, 1, 1.5, 2, 2.5, 3, 3.5, 4, 4.5, 5\}$. As can be seen from the figure, the classification accuracies of all the FL algorithms increase with the increasing $\beta$ value. FedIns, on the other hand, consistently has the highest classification accuracy and is least affected by $\beta$. However, with the increase in inter-client data heterogeneity, the performance of FedAvg~\cite{mcmahan2017communication} will rapidly decline. On the contrary, FedIns is least affected by heterogeneity and still holds an accuracy of $85.13$$\%$ when $\beta$ = $0.1$. Then, we fix $\beta$ = $0.2$, and randomly convert some local images to other styles to control the intra-client data heterogeneity~\cite{geirhos2018imagenet}. 

Similarly, the bigger $\gamma$ values, the higher intra-client data heterogeneity. Fig.~\ref{fig:5} (b) shows the classification accuracies of $\gamma$ values vary in $\{2,4,6,..., 20\}$. Similar to Fig.~\ref{fig:5} (a), the classification accuracy scores of all the FL algorithms decrease with the increasing $\gamma$ value. These baseline methods, in particular, show the most marked downward trend. Nonetheless, when $\gamma$ = $20$, our proposed method only decreases the accuracy from $75.63$$\%$ to ${64.15}$$\%$. In contrast, the classical FL algorithm FedBN, which aims to solve the data heterogeneity problem, decreases the accuracy from $68.53$$\%$ to $\textbf{31.69}$$\%$ when $\gamma$ = $20$.

\subsection{Ablation Study}
\noindent{\textbf{Effectiveness of Core Designs.}}~We first analyze the key components of FedIns on three datasets to evaluate the effectiveness of our core designs. The evaluated components include SSF pool in our model, and SSF, which is the parameter-efficient mechanism in our model.
To this end, we built four derived ablation models, where \emph{Prompt} represents that the SSF of our method is replaced by the similar parameter-efficient mechanism prompts~\cite{jia2022visual}, \emph{Prompt Pool} represents employing the prompt pool mechanism on \emph{Prompt}, \emph{SSF} indicates that the FedIns model removes the SSF pool mechanism, and only retains the parameter-efficient mechanism of SSF in the federated framework, and \emph{SSF Pool}  represents our FedIns that preserves the full components. Following~\cite{jia2022visual}, we add the prompt embeddings with a size of $10\times12\times768$. Table~\ref{tab:2} summarizes the classification accuracy of all ablation models. As compared with \emph{Prompt}, we observed that \emph{SSF} improves the performance from $81.02\%$ to $\textbf{84.32}\%$ on the \textbf{Tiny-ImageNet} dataset, indicating that SSF can provide better performance than prompt since prompt is often sensitive to data and tasks and needs to be carefully designed for each client. Additionally, SSF achieves zero overhead by reparameterizing them into the original pre-trained model weights at the inference stage, while prompt-based methods introduced additional parameters, \ie, \emph{Prompt} with $0.10$ M and \emph{Prompt Pool} with $2.32$ M parameters. This supports our design of using SSF to fine-tune the local model and communicate with the server. \emph{SSF Pool} (\texttt{Ours}) equipped with all components, produces the best classification accuracies, improving the results from $81.02\%$ to $\textbf{86.79}\%$ on the Tiny-ImageNet dataset. 
%On top of that, we compare the inference cost metrics of Prompt/SSF pool as follows: a) Param.: $88.2$/$91.2$M; b) FLOPs: $69.2$/$67.4$G; c) Latency: $54.5$/$80.5$ms.
In general, our core designs can help to alleviate federated data heterogeneity and enhancing the performance of FL algorithms.\label{keycom}

\begin{figure}[t]
    % \vspace{-8pt}
	\begin{center}
		\includegraphics[width=\linewidth]{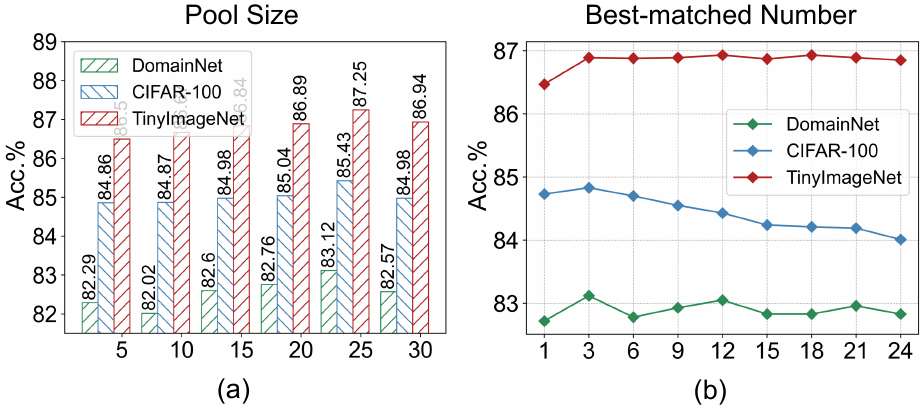}
            \put(-172,3){ \small$M$}
            \put(-50,3.4){ \small$c$}
            % \put(-73,3.4){ \small$\alpha$}
	\end{center}
	\vspace{-15pt}
	\captionsetup{font=small}
    	\caption{\textbf{Ablation studies} versus \textbf{(a)} \textit{Pool size analysis} and \textbf{(b)} \textit{Best-matched number analysis} on \textbf{DomainNet}, \textbf{CIFAR-100}, and \textbf{Tiny-ImageNet} datasets.}
	\vspace{-22pt}
	\label{fig:6}
\end{figure}

% \vspace{2pt}
\noindent{\textbf{Pool Size Analysis.}}~Here, we focus on analyzing the effect of the pool size on model effectiveness. We can freely select the size with the greatest performance even though the number of network parameters increases with pool size because the number of parameters for SSF is small, \ie, $M$ $=$ $1$ only requires $0.20$ M parameters. Additionally, these parameters can be incorporated into the original pre-trained weights by model reparameterization at the inference stage, thereby avoiding additional parameters for the downstream tasks. We record the performance of different pool sizes on the three datasets in Fig.~\ref{fig:6} (a). It can be seen from the figure that as the pool size increases, the performance of FedIns on the three datasets will improve. When $M$ $=$ $25$, our method yields the best classification performance while only requiring $5.35$ M parameters. When $M$ is greater than $25$, the performance degrades due to a large amount of redundancy, which affects the update of the local client by instance-adaptive models. It is worth noting that our model always preserves a higher classification accuracy than the baseline, \ie, when $M = 5$, \texttt{Ours} preserve the results of Acc. $=$ $\textbf{86.50}\%$ $vs.$ FedBN: Acc. $=$ $79.81\%$. 

% \vspace{2pt}
\noindent{\textbf{Best-matched Number Analysis.}}~As the selected SSF values for each input will affect the update of the local model, we need to choose an appropriate best-matched number $C$ to train a good SSF pool. In Fig.~\ref{fig:6} (b), we record the classification accuracy of $c$ values varying in $\{1, 3, 6, ..., 24\}$. As can be seen from this figure, we observe that our proposed method obtains the highest classification accuracy when $C$ $=$ $3$ and the accuracy decreases monotonically on the \textbf{CIFAR-100} when the value of $C$ increases, while remaining invariant on the \textbf{Tiny-ImageNet} dataset. However, even at the lowest point, our method still produces higher classification accuracy than the baselines in the two scenarios, \ie, \texttt{Label Shift} and \texttt{Feature Shift} (see Table~\ref{tab:1}). Therefore, this study reveals the effectiveness of our proposed SSF pool mechanism.

% \noindent{\textbf{Loss Weight Discussion.}}~Here, we investigate the influence of weighting the trade-off between the two loss terms in Eq.~\eqref{eq:0} and report the results on the three datasets in Fig.~\ref{fig:6} (c). The weights of the supervised loss and cosine similarity between query $\boldsymbol{\bm{q}}$ and corresponding matched \texttt{Keys} were determined by the value of $\alpha$. The smaller the value of $\alpha$, the slower the renewal rate of \texttt{Keys}. As can be seen from this figure, our method obtains the best classification results at $\alpha$ $=$ $0.1$. However, when $\alpha$ is higher than $0.3$, the accuracy gradually degrades. This indicates that a small update rate leads to inaccuracy in \texttt{Keys}, thereby destroying the learning of SSF pool.

\vspace{-2pt}

\section{Conclusion}
\vspace{-2pt}
In this work, we re-examine the problem of data heterogeneity in FL, and find that there is not only inter-client data heterogeneity but also intra-client heterogeneity in many complex real-world scenarios, which also significantly degrades FL performance. To address this issue, we propose a new parameter-efficient fine-tuning FL algorithm, FedIns, which makes it possible to use instance-adaptive inference for FL, thus enabling dynamically guided locals to alleviate intra-client data heterogeneity. Unlike existing approaches, FedIns trains an SSF pool for each client and aggregates them into the federated SSF pool on the server. For a given instance, FedIns dynamically finds the best-matched SSF subsets from the pool, and aggregates them to generate an adaptive SSF for instance-adaptive inference. In addition, FedIns introduces only a small number of learnable parameters with the help of the large-scale pre-training model, and greatly reduces the communication cost. Extensive experiments show the superiority of FedIns in handling the inter- and intra-client data heterogeneity issue.

%and then modulates specific SSF for each image input, thereby reducing both the inter- and intra-client data heterogeneity. Additionally, FedIns reduces the amount of local training data with the help of the large-scale pre-training model, and greatly reduces the communication cost and the number of local learnable parameters. Extensive experiments have demonstrated the superiority of FedIns in solving the inter-and intra-client data heterogeneity problem.

\noindent\textbf{Acknowledgements:}
This work was supported by the National Research Foundation, Singapore under its AI Singapore Programme (AISG Award No: AISG2-TC-2021-003) and Agency for Science, Technology and Research (A*STAR) through its RIE2020 Health and Biomedical Sciences (HBMS) Industry Alignment Fund Pre-Positioning (IAF-PP) (grant no. H20C6a0032).

{\small
\bibliographystyle{ieee_fullname}
\bibliography{egbib}
}

%-------------------------
\clearpage
\appendix

\setcounter{table}{0}
\renewcommand{\thetable}{A\arabic{table}}
\setcounter{figure}{0}
\renewcommand{\thefigure}{A\arabic{figure}}

\section*{Contents}
The following items are included in our supplementary material:
\begin{itemize}
  \item Detailed illustration of Fig. {\color{red}1} (c) and (d) in Section~\ref{sec:A}.
  % \item Comparison with the additional baselines in Section~\ref{sec:B}.
  % \item Communication efficiency analysis over the additional baselines in Section~\ref{sec:C}.
  % \item Statistical heterogeneity over the additional baselines in Section~\ref{sec:D}.
  \item Exemplars of the converted style for the intra-client data heterogeneity in Section~\ref{sec:E}.
\end{itemize}

\section{Detailed Illustration of Fig. {\color{red}1} (c) and (d)}\label{sec:A}

\begin{figure}[!ht]
% \vspace{-pt}
	\begin{center}
		\includegraphics[width=\linewidth]{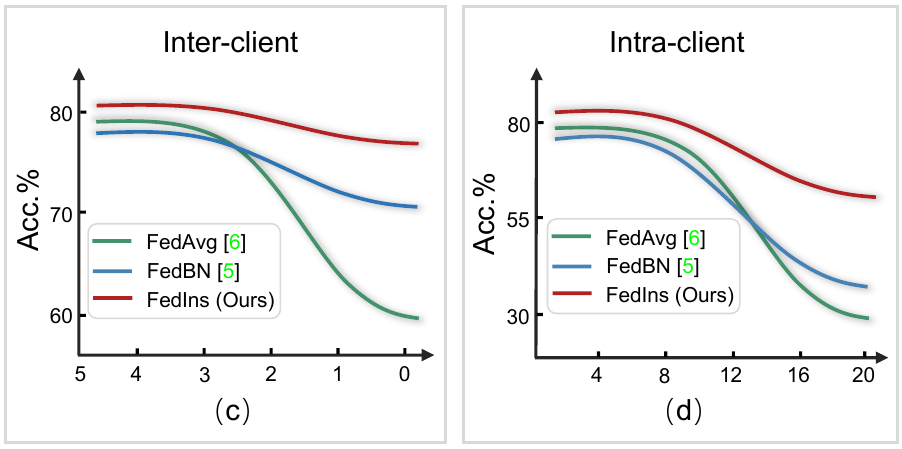}
        \put(-170,7){ \small$\beta$}
        \put(-50,9){ \small$\gamma$}
        \end{center}
	\captionsetup{font=small}
       \vspace{-7pt}
	\caption{\small\textbf{Accuracy comparisons} of FedAvg~\cite{mcmahan2017communication}, FedBN~\cite{li2021fedbn}, and FedIns in terms of various \textbf{(a)} \textit{inter-client heterogeneity} and \textbf{(b)} \textit{intra-client heterogeneity} on \textbf{DomainNet}, where $\beta$ is hyperparameter of Dirichlet distribution, the smaller $\beta$ values, the higher inter-client heterogeneity; $\gamma$ is the number of the converted styles (see Sec.~\ref{sec:E} for details), the larger $\gamma$ values, the higher intra-client heterogeneity.}
	\label{fig:A1}
 \vspace{-10pt}
\end{figure}

We redisplay Fig. {\color{red}1} (c) and (d) in Fig.~\ref{fig:A1} in details. As can be seen from this figure, the accuracy \% comparisons of FedAvg~\cite{mcmahan2017communication}, FedBN~\cite{li2021fedbn}, and FedIns in terms of various \textbf{(a)} \textit{inter-client heterogeneity} and \textbf{(b)} \textit{intra-client heterogeneity} on \textbf{DomainNet} are recorded. We use the hyperparameter $\beta$ of Dirichlet distribution to control the inter-client heterogeneity and $\gamma$ to control the intra-client heterogeneity. The smaller $\beta$ values, the higher inter-client heterogeneity; the larger $\gamma$ values, the higher intra-client heterogeneity, where $\gamma$ is the number of the converted styles~\cite{geirhos2018imagenet}. We provide the exemplars of the converted styles in Sec.~\ref{sec:E}.

\section{Exemplars of the Intra-client Data Heterogeneity}\label{sec:E}
\begin{figure*}[!t]
    % \vspace{-2pt}
	\begin{center}
		\includegraphics[width=\linewidth]{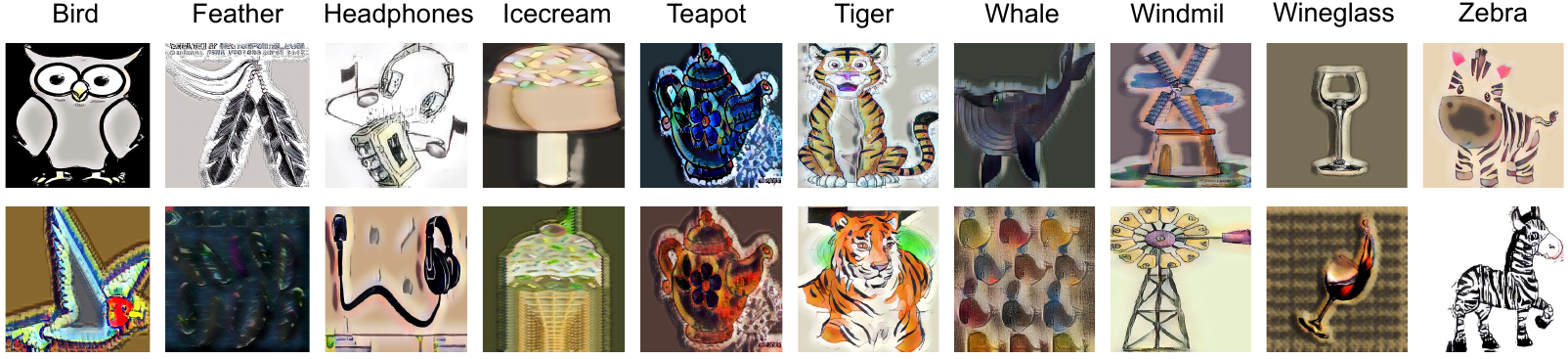}
	\end{center}
    \vspace{-8pt}
	\captionsetup{font=small}
    	\caption{\textbf{Visualization} of the convert images with various style on \textbf{DomainNet} to control the intra-client data heterogeneity. }
	\vspace{-3pt}
	\label{fig:A4}
\end{figure*}

As we mentioned, we randomly convert some local images to other styles to control the intra-client data heterogeneity~\cite{geirhos2018imagenet}. Here, we show some exemplars of the converted images on the client \texttt{clipart} under \textbf{DomainNet} dataset in Figure.~\ref{fig:A4}. As can be seen from this figure, different styles show great differences, thereby making the local data preserve higher intra-client data heterogeneity and being able to be used to demonstrate the effectiveness of our method.

% {\small
% \bibliographystyle{ieee_fullname}
% \bibliography{egbib}
% }

% {\small
% \bibliographystyle{ieee_fullname}
% \bibliography{egbib}
% }

\end{document}